\documentclass[compactaffiliation]{Interspeech}

\usepackage{hyperref}
\usepackage[numbers]{natbib}
\interspeechcameraready

\title{Overcoming Data Scarcity in Multi-Dialectal Arabic ASR via Whisper Fine-Tuning}

\author[affiliation={1, 2}]{Ömer Tarik}{Özyilmaz}
\author[affiliation={3}]{Matt}{Coler}
\author[affiliation={2}]{Matias}{Valdenegro-Toro}

\affiliation{Department of Internal Medicine, Division of Nephrology}{University of Groningen, University Medical Center Groningen}{The Netherlands}
\affiliation{Bernoulli Institute for Mathematics, Computer Science, and Artificial Intelligence}{University of Groningen}{The Netherlands}
\affiliation{Speech Technology Lab, Campus Fryslân}{University of Groningen}{The Netherlands}
\email{o.t.ozyilmaz@umcg.nl, m.coler@rug.nl, m.a.valdenegro.toro@rug.nl}

\usepackage{comment}

\begin{document}

\maketitle

\begin{abstract}
    
    \noindent Although commercial Arabic automatic speech recognition (ASR) systems support Modern Standard Arabic (MSA), they struggle with dialectal speech. We investigate the effect of fine-tuning OpenAI's Whisper on five major Arabic dialects (Gulf, Levantine, Iraqi, Egyptian, Maghrebi) using Mozilla Common Voice for MSA and the MASC dataset for dialectal speech. We evaluate MSA training size effects, benefits of pre-training on MSA data, and dialect-specific versus dialect-pooled models. We find that small amounts of MSA fine-tuning data yield substantial improvements for smaller models, matching larger non-fine-tuned models. While MSA pre-training shows minimal benefit, suggesting limited shared features between MSA and dialects, our dialect-pooled models perform comparably to dialect-specific ones. This indicates that pooling dialectal data, when properly balanced, can help address data scarcity in low-resource ASR without significant performance loss.
\end{abstract}

\section{Introduction}
Automatic Speech Recognition (ASR) systems have achieved remarkable performance for major world languages, yet they often fail to serve speakers of regional dialects and variants \cite{alsayadi_deep_2022}. This disparity is particularly evident in Arabic, where commercial ASR systems primarily support Modern Standard Arabic (MSA) --- the formal, standardized variant used in broadcast media and official communication --- while struggling with the rich diversity of regional dialects spoken in daily life. This gap impacts millions of Arabic speakers who rely on their regional
dialects for daily communication \citep{vergyri2004automatic}. The lack of robust dialectal ASR systems creates significant barriers to technology access to those who primarily use their regional dialects in daily life. Without effective dialectal support, users are often forced to code-switch to MSA when using voice assistants, creating an unnatural and potentially exclusionary interaction. This affects applications from healthcare to business where accurate transcription is essential.

Despite Arabic's large speaker population, its dialects face many of the same challenges as typical low-resource languages in ASR development. In their review, \citet{alsayadi_deep_2022} identified several critical barriers: training data for Arabic ASR remains skewed towards MSA, dialectal datasets lack volume and variation, and commercial systems that claim multi-dialect support show significant performance degradation when handling regional variants. These challenges are acute in real-world applications, where ASR systems trained on MSA struggle to generalize across the phonological and lexical variations present in dialectal speech. Current models show inconsistent performance across dialects and datasets \cite{abdelali_larabench_2024}, further complicated by training data imbalances. This imbalance risks creating ASR systems that exhibit systematic bias, performing well for majority dialects while marginalizing speakers of less-resourced variants.

\noindent These challenges highlight two critical requirements for advancing dialectal Arabic ASR: developing methods that perform robustly with limited and homogeneous training data, and creating unified models that maintain accuracy across multiple dialects despite dataset imbalances. While large multilingual models like Whisper \cite{radford_robust_2023} and USM \cite{zhang_google_2023} show promise for low-resource languages, they trade broad coverage for per-language accuracy. Additionally, the substantial computational requirements of these advanced, multilingual models constrain their deployment on devices with limited resources, as optimizing them for such environments could potentially degrade performance for less commonly spoken languages. We investigate whether small models with fine-tuning could achieve similar performance to larger models. Next to this, this paper will focus on leveraging the advantages of both large ASR models and using the large amount of MSA speech data to develop a novel dialectal Arabic ASR system and investigate its strengths and weaknesses. This includes the linguistic diversity across dialects as well as different combinations and sizes of training data. 

% The following paper will investigate the effect of fine-tuning Whisper first with MSA (pre-training) and subsequently with the Gulf-, Levantine-, Iraqi-, Egyptian-, and Maghrebi-Arabic dialects. The nature and extent of the impact of fine-tuning will then be investigated for the different dialects and across different pre-training configurations. This includes the linguistic diversity across dialects as well as different types and sizes of training data. 

\noindent This study investigates three key research questions: (1) What is the effect of MSA training data size on ASR performance? (2) Is MSA pre-training beneficial for dialectal Arabic fine-tuning? (3) How do dialect-specific and dialect-pooled models compare in performance?

\noindent We hypothesize that MSA pre-training will enhance dialectal ASR performance through shared linguistic features \cite{yu2010roles}, with improvements varying by dialect and training configuration. We expect diminishing returns with increased training data and superior performance from dialect-specific models over pooled models \cite{alsayadi_deep_2022}.

\section{Related Work}

\citet{alsharhan_investigating_2020} demonstrated that dialect-specific training outperformed multi-dialect approaches, finding particular divergence in Maghrebi Arabic. Using Deep Neural Network Hidden Markov Models (DNN-HMM), they showed that increased training data in general models did not necessarily improve performance over specialized ones. However, their findings for traditional hybrid models may not extend to end-to-end approaches. Recent end-to-end models demonstrated strong multilingual performance through general speech representation learning and multi-task training \cite{radford_robust_2023, zhang_google_2023}. The fine-tuning of networks such as Whisper by \citet{radford_robust_2023} and USM by \citet{zhang_google_2023} was compared against state-of-the-art systems by \citet{abdelali_larabench_2024} in a novel Arabic benchmark. They demonstrated that their specialized state-of-the-art system outperformed multilingual zero-shot networks such as USM and Whisper in most Arabic speech datasets, with USM showing the edge in some cases. They mentioned that even though USM outperforms Whisper in all cases, fine-tuning Whisper on just 2 hours of speech data drastically improved this and came near to closing this gap. Unfortunately, \citet{abdelali_larabench_2024} omitted fine-tuning Whisper on larger amounts of data or dialect-specific data. The advantage of Whisper is the open-source nature compared to the closed-source, more powerful USM, which allows Whisper to be fine-tuned and compared against domain-specific models.

Research on both dialectal diversity using DNNs and fine-tuning multilingual ASR systems on larger amounts of data in the target domain is present but limited to smaller dialects \cite{rijal2024whisper, torgbi2025adapting}. The current paper investigates the effect and extent of shared feature learning from MSA in the context of dialectal Arabic ASR performance and the level of linguistic similarity between dialects according to a DNN architecture is presented.

\section{Methodology}
\subsection{Data}
% explain datasets + preprocessing
\textbf{Mozilla Common Voice} is a widely favored option in the ASR industry for accessing both high-quality and extensive datasets \cite{commonvoice}. The dataset was collected through volunteers writing, recording, and/or validating samples. We used the Arabic partition of Common Voice 16.1\footnote{Mozilla Common Voice 16.1 can be found at \url{https://commonvoice.mozilla.org/en/datasets}}. This partition contained mostly MSA samples, as the contributed sentences are expected to be, which is why we can treat it as our MSA dataset. 

The audio was resampled from 48 kHz to 16 kHz to accommodate the feature extractor of the Whisper model architecture. Afterwards, the log-Mel features were extracted from the audio samples. The labels were obtained by tokenizing the transcriptions with the Whisper tokenizer and applying zero-padding such that all labels had the same fixed dimension. A split already existed in the original HuggingFace implementation of the dataset. The original ``train'' and ``validation'' partitions were merged, totaling 40 hours of speech, then reshuffled with a fixed random seed, and finally split again with a fixed random seed with a 80:20 ratio into a train- and validation-set, respectively. The ``test'' partition contained 13 hours of speech and was kept separate until final evaluation.

The \textbf{Massive Arabic Speech Corpus (MASC)} dataset is another large-scale Arabic ASR dataset \cite{masc}. The data was crawled from YouTube videos, consists of over 1,000 hours of speech and required segmentation from full audio files and matching with corresponding subtitle time-segments. Each processed sample combined four elements: the transcribed text, segment duration, dialect classification (defined in the metadata), and the audio segment itself. The dialects of interest were the five large dialects utilized by \citet{alsharhan_investigating_2020}, namely Gulf, Levantine, Iraqi, Egyptian, and Maghrebi. Since the MASC audio was already at 16 kHz, no resampling was required. Finally, we extracted log-Mel features and applied the Whisper tokenizer with zero-padding similar to the Common Voice dataset.

The MASC dataset also contained a train-, validation-, and test-set. We decided to use the training set only since the size was already sufficient. A large imbalance was present in the dataset, as the Egyptian (385h) and Levantine (148h) datasets were clearly overrepresented, while Iraqi (13h) and Maghrebi (17h) were more limited. Due to this imbalance, a maximum of 20 hours of speech per dialect was used. Dialects with more than 20 hours were randomly shuffled and undersampled, while dialects with less than 20 hours are used in full. 

\noindent From these resampled datasets, we split the data with a 80:20 ratio into a train- and test-set with a fixed random seed. These test sets are used for final evaluation. The train partition is subsequently split further into a 80:20 partition of train and validation, respectively, again with a fixed random seed. Finally, all datasets are filtered on transcriptions that are longer than 448 tokens, because of the model configuration. This filtering only removes two samples from the Egyptian train set and one sample from the Maghrebi test set. The datasets are ported into the HuggingFace \texttt{Dataset} format, making the data-loading identical to the Common Voice data.

\subsection{Model}\label{sec:model}
% explain whisper
The model of choice, as explained previously, is the state-of-the-art and multilingual ASR system Whisper by OpenAI \cite{radford_robust_2023}. The reason for using Whisper is its open-sourced nature compared to USM \cite{zhang_google_2023}, and its impressive multilingual performance. The best model for this comparison was deemed to be the \texttt{whisper-small} configuration, as its size is located amidst two smaller and two larger versions. Furthermore, this configuration still has a competitive inference speed on lower-grade hardware, while containing enough parameters to exhibit model flexibility. For details, the reader is directed to \citet{radford_robust_2023}. The weights are obtained through the HuggingFace Transformers library \cite{wolf2019huggingface}, using the \texttt{openai/whisper-small} (244 million parameters) checkpoint.

For each of the five dialects and a dialect-pooled scenario (trained on all dialects), three fixed random seeds were used for the train-validation split and each of these were trained with- and without pre-training on MSA. Six scenarios and six models per scenario resulted in 36 models. Furthermore, the \texttt{whisper-small} checkpoint was fine-tuned with 20\%, 40\%, 60\%, 80\%, and 100\% of the MSA data, to obtain a course of the effect of the amount of data. The data subsets are obtained by taking a random subset of the data with a fixed random seed. This results in five additional models being trained. These models were evaluated with three different fixed random seeds (42, 84, and 168) by taking a random 80\% of the MSA test set. The reported WER and CER results for each model represent the average performance across these three evaluations.

All models were trained for a maximum of 6,000 steps with a batch size of 8. Early stopping was used to halt training if no decrease in evaluation loss was observed after three evaluation rounds. Thirty-three out of the total 36 models triggered early stopping, at an average of 2,682 steps. For each model, the checkpoint with the lowest validation loss was selected as the final model for evaluation. AdamW was used as an optimizer with a linear learning rate scheduler \cite{kingma2014adam, adamw}. The learning rate increased from $5.0 \times 10^{-7}$ until the maximum learning rate of $1.0 \times 10^{-5}$ in the first 500 (warm-up) steps, after which it decreased again. The hyperparameters $\beta_1$ and $\beta_2$ of AdamW were kept at $0.9$ and $0.99$, respectively. Since this is a fine-tuning task, the learning rate was kept as low as possible to not disturb the original weights of the Whisper model. Evaluation was performed every 250 steps, after which a model checkpoint was saved as well. The loss function used was cross-entropy.

\subsection{Tools and Technologies}

The fine-tuning and evaluation of the models were performed by an NVIDIA A100 GPU. The implementation was done using Python version 3.10.4, PyTorch version 1.12.1 \cite{pytorch}, and HuggingFace Transformers version 4.39.3 \cite{wolf2019huggingface}. The code is available on GitHub at \url{https://github.com/O-T-O-Z/finetune-ar-dialects}. All additional requirements and dependencies can be found there as well. The fine-tuned models and pre-processed datasets are available on HuggingFace at \url{https://huggingface.co/collections/otozz/arabic-dialects-67949653e522c7de7fdddc7a}. The Whisper fine-tuning blog post by Sanchit Gandhi from HuggingFace was used as inspiration \cite{hftutorial}.

% \subsection{Ethical considerations}
% \label{ethics}
% Both datasets that were used are available under a Creative Commons license. Participants consented to their data being used for research purposes, and can opt out at any time. Furthermore, the participants are anonymized. Finally, the results are presented as fairly and honestly as possible, with possible mistakes or omissions being out of the hands of the researchers.

\subsection{Metrics}\label{sec:metrics}
The metrics of choice are the word error rate (WER) and character error rate (CER). WER is defined as the ratio between the number of modifications or edits and the total number of words:

\begin{equation}
    \text{WER} = \frac{S_{w}+I_{w}+D_{w}}{N_{w}},
    \label{eq:wer}
\end{equation}

\noindent where $S_{w}$ denotes the number of substitutions, $I_{w}$ the number of insertions, $D_{w}$ the number of deletions, and $N_{w}$ the number of words in the reference or ground truth. A WER can exceed 100\% when more words are inserted, deleted, and/or substituted than there were words in the reference, which is often an indication of ``hallucination'', or the model generating nonsensical and exceedingly long output. The lowest WERs for English ASR are currently below 10\% \cite{huggingface_open_asr_leaderboard}, but many other languages are higher. The CER is defined as the ratio between the number of edits and total number of characters in the ground truth and is also evaluated to detect any inconsistencies in analysis.

\section{Results and Discussion}

\subsection{Effect of MSA training size}\label{sec:resex1}
\begin{figure}[!ht]
    \centering
    \includegraphics[scale=0.3]{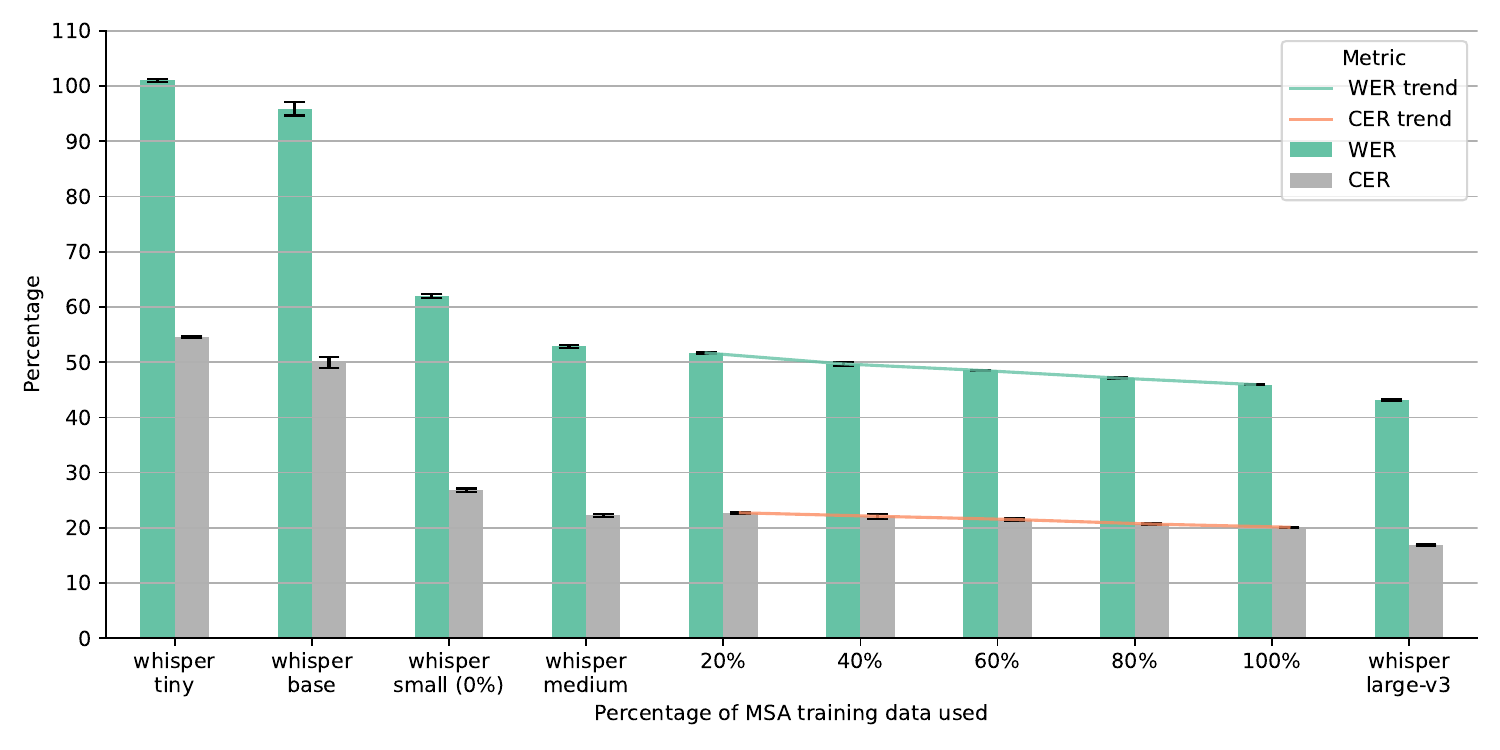}
    \caption{Barplot shows increasing train size for \texttt{whisper-small} results in better performance that nears larger model performance.}
    \label{fig:ex_trainsize}
\end{figure}

The goal of the first experiment is to visualize the effect of the MSA training data size on the performance of Whisper. As observed in Figure \ref{fig:ex_trainsize}, the expected downward trend in both WER and CER was recognized. This decrease was most noticeable when moving from \texttt{whisper-small} to the model fine-tuned on just 20\% of the MSA training dataset. There was an improvement of almost 10\% visible in WER. After that, another 80\% was required to improve the model by around 6\%. This validates our hypothesis and leads us to believe that even small amounts of training data can be extremely beneficial for the performance of Whisper on MSA. Furthermore, our findings are largely in line with \citet{alsharhan_investigating_2020}, who found an increase in model performance with more data up to a certain extent. Even though they performed this similar experiment with multi-dialectal data, the conclusion is equivalent.

The most interesting outcome of this experiment was the comparable performance of the fine-tuned \texttt{whisper-small} model checkpoint with the much larger \texttt{whisper-large-v3} checkpoint (1550 million parameters) in the ASR performance of MSA. Even though the latter has access to higher model complexity, the former was able to specialize in the specific task of ASR of MSA with fewer weights and a limited amount of fine-tuning. The average inference time per sample was measured as 0.12s for \texttt{small} and 0.29s for \texttt{large-v3} (for the MSA test set). Inference time and computational cost can thus be reduced heavily through a small drop in WER performance \cite{radford_robust_2023}.

\subsection{Impact of Pre-training}\label{sec:resex2}
\begin{figure}[!ht]
    \centering
    \includegraphics[scale=0.2]{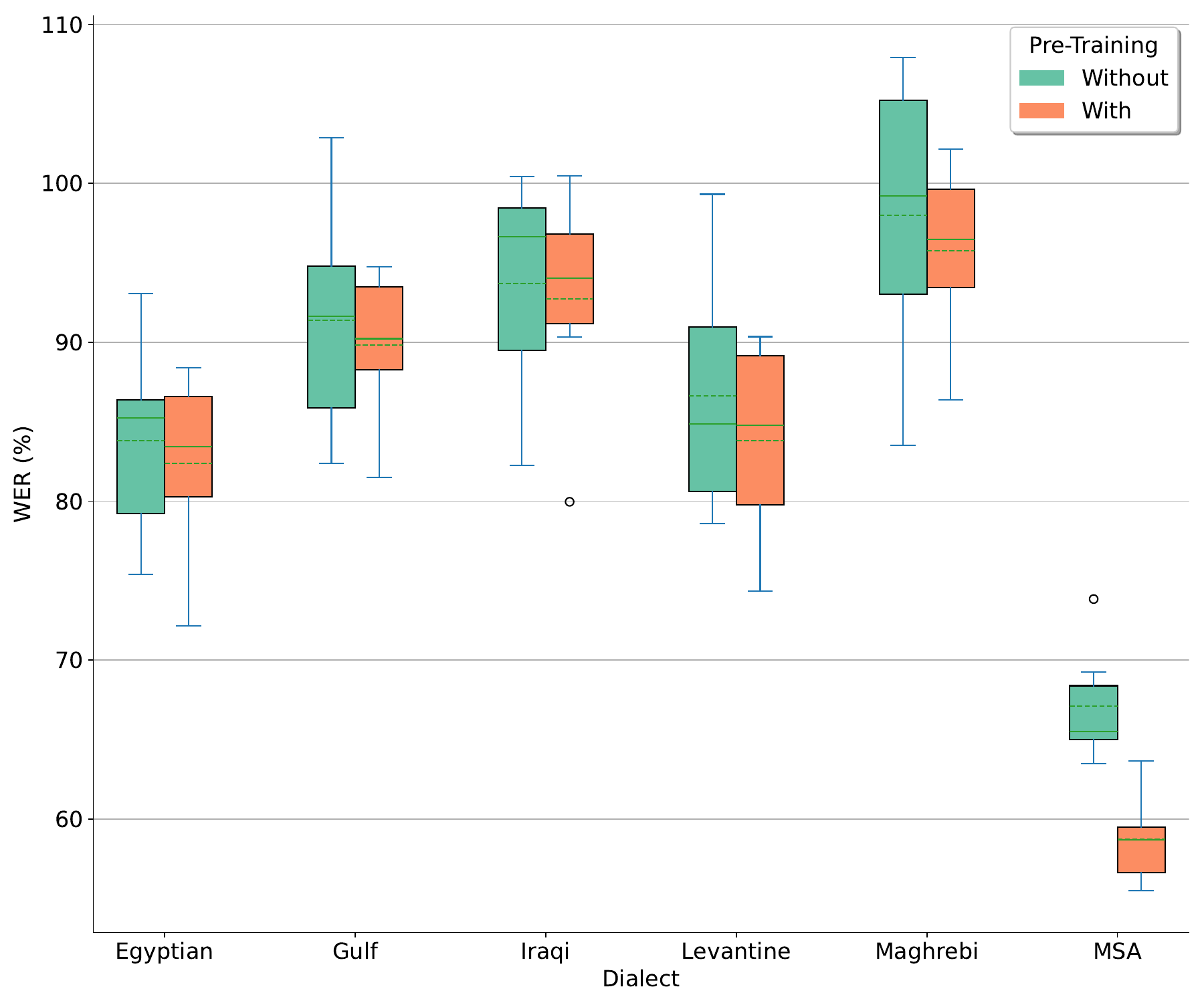}
    \caption{Differences are small between pre-training and no pre-training, except for MSA.}
    \label{fig:ex_ttest_within}
\end{figure}

The results of the remaining 36 models are found in Figure \ref{fig:ex_ttest_within}. The models with- and without pre-training are compared on the test sets in Figure \ref{fig:ex_ttest_within}. No large differences between pre-training and no pre-training was observed, as most results were close to each other in terms of WER. The sole exception was the MSA test set, where a large difference was visible. To validate our findings, multiple Wilcoxon signed-rank tests were performed, each on the differences between test set performances. The Holm-Bonferroni correction was applied to correct for multiple comparisons and no differences were found between the test set performances. For MSA, a slight drop in performance was definitely present, since Figure \ref{fig:ex_trainsize} showed a $45.93\%$ WER on the pre-trained model before fine-tuning, while $\mu=58.71$ for all pre-trained models after fine-tuning. This is likely due to the model ``forgetting'' or re-calibrating its weights to try to accommodate both the new information and the previously learnt information.

Furthermore, another Wilcoxon signed-rank test was performed to investigate the difference between the two groups of models as a whole. Pre-trained models ($\mu=83.87$, $\sigma=13.24$) showed better performance ($Z=120.00$, $p=5.1\mathrm{E}{-4}$) than non pre-trained models ($\mu=86.77$, $\sigma=11.94$). This could indicate that some general information about Arabic is preserved and generally beneficial for performance.

% We hypothesized a more pronounced improvement because we expected the pre-trained network to generalise and identify linguistic features that are common across all variations of Arabic better than Whisper without pre-training. One possible reason could be that the variations between dialects and between each dialect and MSA are too vast to generalise. This could be due to the limited number of common words or sentence structures between the two. Alternatively, since Whisper has already been trained on multiple languages, this type of information could already be present in the network. Finally, we used the smaller checkpoint of Whisper, which has limited capacity to learn compared to the larger sizes.

\subsection{Dialect-Specific vs Pooled Models}\label{sec:resex3}
\begin{figure}[!ht]
    \centering
    \includegraphics[scale=0.2]{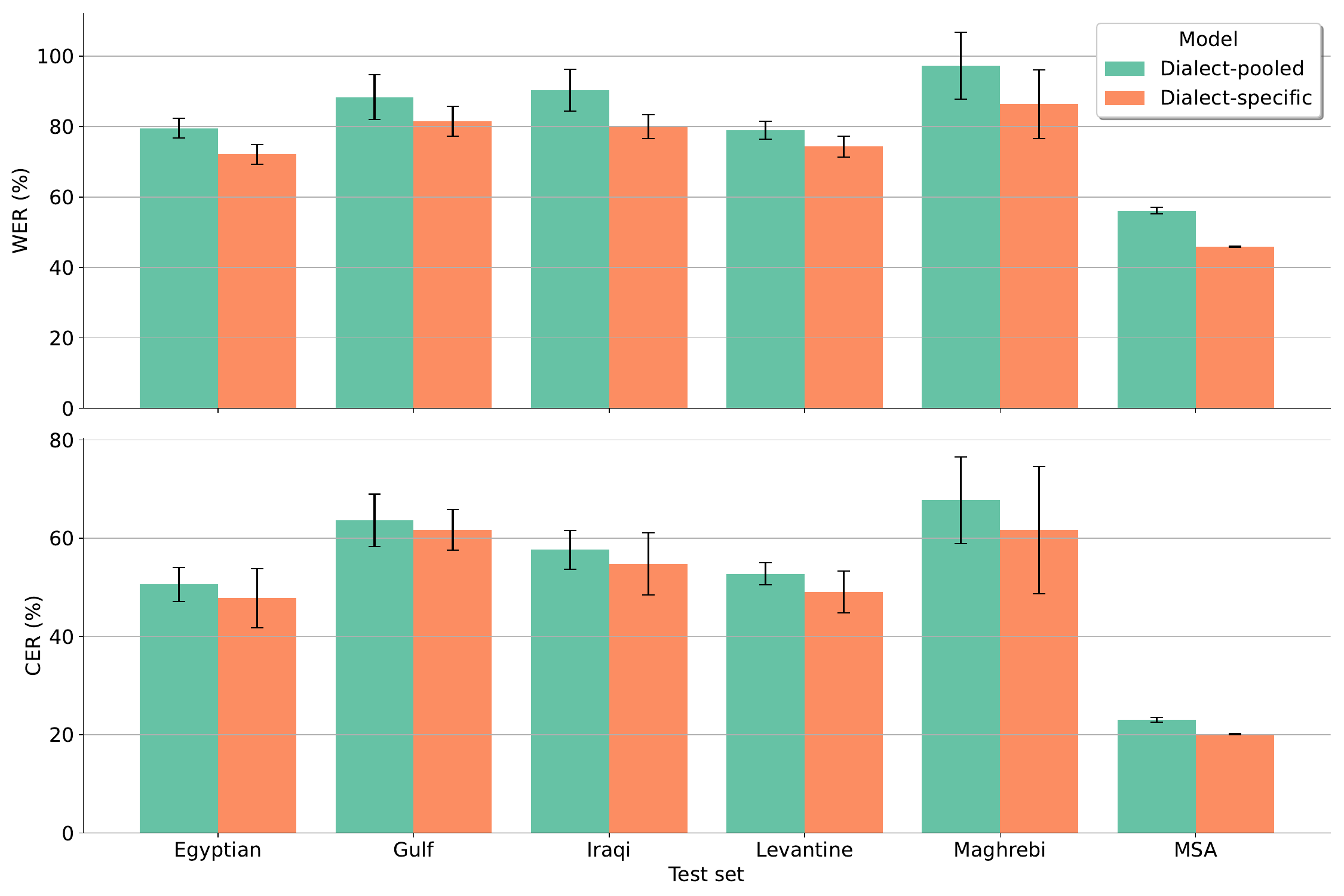}
    \caption{Dialect-pooled and dialect-specific models perform close to each other on all dialectal test sets (both for WER and CER).}
    \label{fig:ex_comparison_pooled}
\end{figure}

For the difference in performance between the dialect-specific and dialect-pooled models, we focused on the performance of the pre-trained models per dialectal training set. Models trained on a specific dialectal dataset generally worked best on the MSA test set (likely due to pre-training), followed by their corresponding test set. The model fine-tuned on the dialect-pooled training set performed close to the models fine-tuned on each dialect specifically (Figure \ref{fig:ex_comparison_pooled}). This is contrary to previous literature that mention dialect-pooled performance as a challenge \cite{alsayadi_deep_2022}. The average drop in WER performance from dialect-pooled to dialect-specific models is $8.42\%$, while CER performance reduces by $3.37\%$, which indicates that only a small amount of performance is sacrificed when opting for a dialect-pooled approach. Better yet, when considering the improved performance on the other dialects that could be considered out-of-distribution for the dialect-specific models, the dialect-pooled models show much potential in overall performance. 
% These results are also in line with \citet{alsharhan_investigating_2020}, who found a difference of roughly $2\%$. Thus, we have reason to believe that the challenges of insufficiency and lack of variation can be reduced through pooling the dialectal datasets. The fear of introducing of imbalance can be easily mitigated by ensuring data balancing before training. Even the difficulties imposed by the low-resource nature of the dialectal datasets can already be reduced by using a dialectal dataset of only 20 hours. Thus, by starting with a pre-trained ASR system such as Whisper, one reduces the data requirements significantly. 

\begin{figure}[!ht]
    \centering
    \includegraphics[scale=0.43]{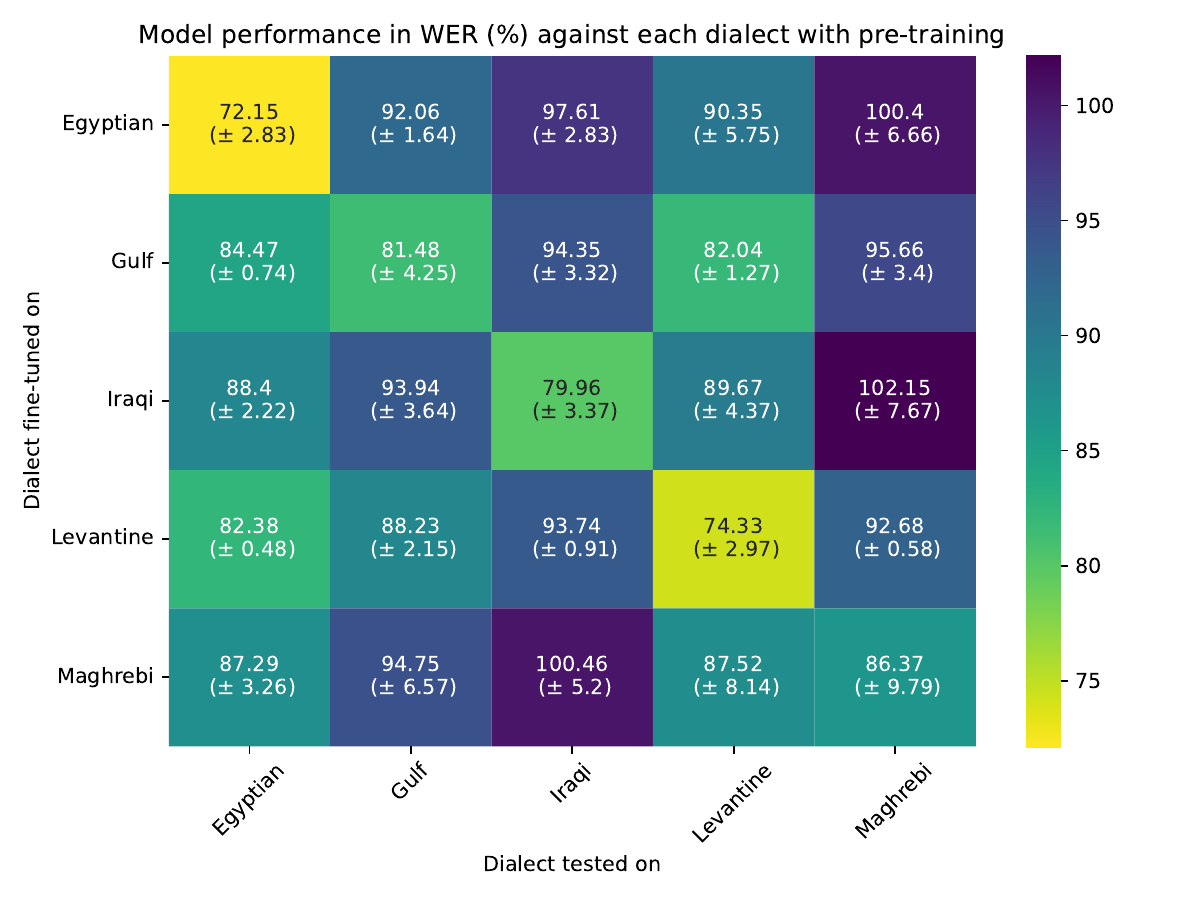}
    \caption{Confusion matrix of WER performance between the dialectal train and test sets. Iraqi and Maghrebi show clear distance from other sets.}
    \label{fig:ex_finetune_cm_wer}
\end{figure}

\noindent Another advantage of using the dialectal speech datasets for training, is the ability it provided us to investigate the linguistic differences, similar to \citet{alsharhan_investigating_2020}. In Figure \ref{fig:ex_finetune_cm_wer}, we find that fine-tuning on Egyptian, Iraqi, and Maghrebi datasets decreases performance on other dialects most (rows). This is contrary to \citet{alsharhan_investigating_2020}, who reported that Egyptian was similar to other Arabic dialects. The authors also note that Maghrebi is farthest from other dialects due to French influence on the language, which was observed in the poor performance on the Maghrebi test set when trained on a different dialect.
Further, the large difference between Iraqi and Maghrebi was arguably expected due to the large geographical distance between the two regions. Conversely, Levantine and Gulf are close both geographically and in terms of cross-dialect performance. These results cannot be taken as a truth and a linguistic study would obviously be more suited for this purpose. Yet, the results for end-to-end models are mostly in line with similar comparisons for DNN-HMM models and invite further exploration in linguistic studies \cite{alsharhan_investigating_2020}.

\subsection{Limitations}

A technical limitation was our reliance on Whisper's built-in tokenizer configured for Arabic transcription, without implementing additional dialect-specific text normalization. Dialectal Arabic benefits from specialized text normalization including standardization of character variants (for example, different forms of alef and hamza), diacritic handling, and normalization of dialectal spelling variations. This limitation may have affected our WER measurements across dialects, as dialectal variations in spelling and character usage could inflate error rates even when the phonetic content was correctly recognized. Future work should investigate the impact of dialect-aware text normalization on ASR performance.

\section{Conclusion}
Data scarcity poses a problem in dialectal Arabic ASR, as most commercial applications are focused on MSA performance. All in all, we demonstrated that fine-tuning \texttt{whisper-small} with a small amount of MSA data can achieve performance comparable to \texttt{whisper-large-v3} at a fraction of the inference time. This could have a profound impact on industry-grade applications of (Arabic) ASR, with the deployment of smaller models being preferred over large, computationally expensive, and slow models. Costs will be reduced on the side of the producer and the response time decreases on the client's side. Pre-training on MSA was also found to be beneficial for average performance in a downstream dialectal fine-tuning task, which means that training time can be reduced as well.

Further optimization showed that it seems beneficial to combine dialectal datasets for a more robust DNN due to the larger size and increased variation of its training dataset. This could be proposed as a solution to the lack of variation and data present in Arabic ASR as well as low-resource language ASR in general. By pooling similar dialects, one could reduce computational burden. When there are more similarities than in dialectal Arabic, generalization is expected to increase even further, such as with cross-continental Spanish or Portuguese.

Finally, large performance differences were present between the Arabic dialect-specific models. Dialects that are geographically further apart showed larger differences, such as the Maghrebi (North-African) and Iraqi dialects, while regions that are closer to each other showed more similar performances, for instance the Levantine (Jordan, Palestine) and Gulf (UAE, Saudi Arabia) dialects.

\ifinterspeechfinal
\section{Acknowledgements}
% Acknowledgement should only be included in the camera-ready version, not in the version submitted for review. The 5th page is reserved exclusively for acknowledgements and  references. No other content must appear on the 5th page. Appendices, if any, must be within the first 4 pages. The acknowledgments and references may start on an earlier page, if there is space.
% The authors would like to thank ISCA and the organising committees of past Interspeech conferences for their help and for kindly providing the previous version of this template.
The authors would like to thank Mo Assaf for the inspiration, and Nikolai Herrmann, Ivo P. de Jong, and dr. Marco Zullich for their input in the final phases of the paper. 
\else
\fi
\section{References}
{\def\section*#1{}
\bibliographystyle{IEEEtranN}
\bibliography{mybib}
}
\renewcommand{\refname}{}

\end{document}